\documentclass[10pt,twocolumn,letterpaper]{article}

\usepackage{cvpr}
\usepackage{times}
\usepackage{epsfig}
\usepackage{graphicx}
\usepackage{amsmath}
\usepackage{amssymb}

\usepackage{amsmath} 
\usepackage{tabularx}
\usepackage{diagbox}
\usepackage{booktabs} 
\usepackage{algorithm}  
\usepackage{algorithmicx}  
\usepackage{algpseudocode}  
\usepackage{amsmath} 
\floatname{algorithm}{Algorithm}

\cvprfinalcopy 

\def\cvprPaperID{4666} 

\begin{document}
	
	\title{Fully Learnable Group Convolution for Acceleration of Deep Neural Networks}
	
	\author{Xijun Wang$^{1,2}$ \qquad  Meina Kan$^1$ \qquad Shiguang Shan$^{1,2,3}$ \qquad Xilin Chen$^{1,2}$\\
		$^1$Key Lab of Intelligent Information Processing of Chinese Academy of Sciences (CAS),\\ Institute of Computing Technology, CAS, Beijing 100190, China\\
		$^2$University of Chinese Academy of Sciences, Beijing 100049, China\\
		$^3$CAS Center for Excellence in Brain Science and Intelligence
		Technology, Shanghai 200031, China\\
		{\tt\small xijun.wang@vipl.ict.ac.cn \qquad \{kanmeina,sgshan,xlchen\}@ict.ac.cn}
	}

	\maketitle
	
	\begin{abstract}
		Benefitted from its great success on many tasks, deep learning is increasingly used on low-computational-cost devices, e.g. smartphone, embedded devices, etc. To reduce the high computational and memory cost, 
		in this work, we propose a fully learnable group convolution module (FLGC for short) which is quite efficient and can be embedded into any deep neural networks for acceleration. Specifically, our proposed method automatically learns the group structure in the training stage in a fully end-to-end manner, leading to a better structure than the existing pre-defined, two-steps, or iterative strategies. Moreover, our method can be further combined with depthwise separable convolution, resulting in 5$\times$ acceleration than the vanilla Resnet50 on single CPU. An additional advantage is that in our FLGC the number of groups can be set as any value, but not necessarily $2^k$ as in most existing methods, meaning better tradeoff between accuracy and speed. As evaluated in our experiments, our method achieves better performance than existing learnable group convolution and standard group convolution when using the same number of groups.
	\end{abstract}
	
	\section{Introduction}
	\begin{figure*}[t]
		\begin{center}
			\includegraphics[width=1\linewidth]{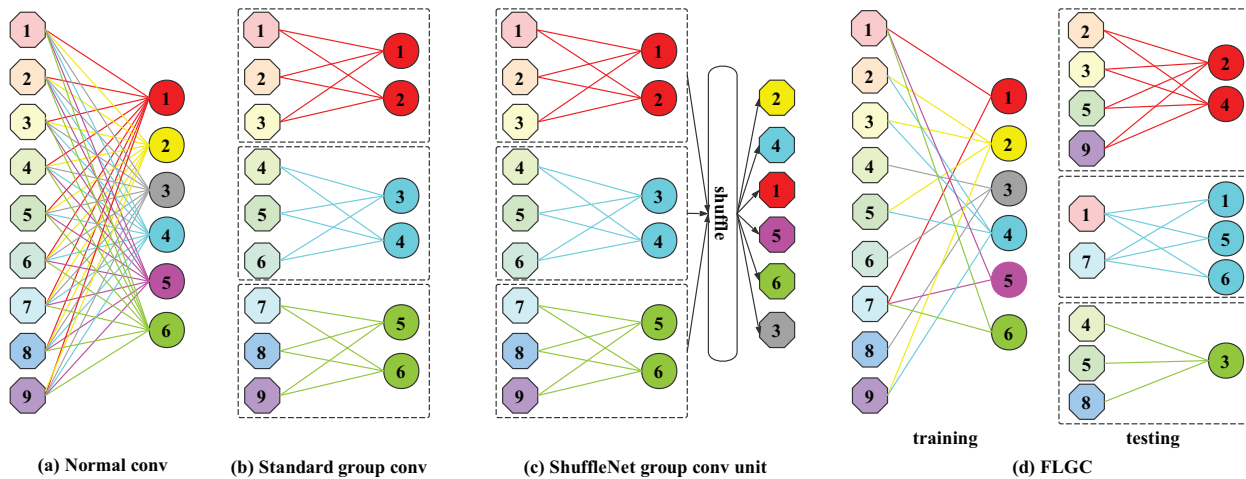}
		\end{center}
		\caption{An overview of different group convolution mechanisms. (a) is a normal convolution. (b) is a standard group convolution, in which input channels and filters in each group are both fixed. (c) is ShuffleNet group convolution unit, in which input channels are fixed. (d) is our FLGC convolution, in which the grouping structure including both input channels and filters in each group is dynamically learnt. The octagons represent the input channels and the circles represent the filters.}
		\label{fig:fig1}
		\vspace{-10pt}
	\end{figure*}
	Since the Alexnet proposed by Krizhevsky \etal~\cite{krizhevsky2012imagenet} achieved breakthrough results in the 2012 ImageNet Challenge, deeper and larger convolutional neural networks (CNNs) have become a ubiquitous setting for better performance, especially on tasks with big data~\cite{deng2009imagenet, lin2014microsoft}. However, even an ordinary CNNs is usually with dozens, hundreds or even thousands of layers and thousands of channels ~\cite{he2016deep, szegedy2017inception, huang2017densely}. Such huge parameters and high computational cost make it insupportable on devices with limited hardware resources or applications with strict latency requirements. In \cite{denil2013predicting}, Misha Denil \etal found that there is significant redundancy in CNNs, and the accuracy will not drop even many of the network parameters are not learned or removed. After that, various methods of reducing redundancy have emerged. These methods can be roughly grouped into two categories, post-processing methods such as pruning or quantizing a pre-trained deep model, and efficient architecture design methods attempting to design fast and compact deep network. 
	
	\subsection{Post-processing methods}
	A straightforward strategy is to post-process a pre-trained model, such as pruning the parameters, or quantizing the model by using fewer bits. 
	
	\textbf{Parameter Pruning.} 
	Some \textit{ fine-grained methods} attempt to prune the wispy  connections between two neural nodes based on its importance, and thus convert a dense network to a sparse one~\cite{liu2015sparse, han2015learning, guo2016dynamic, lebedev2016fast}. A typical one is \cite{han2015learning}, in which Han \etal~proposed to learn the importance of each connection and then those unimportant connections are removed to reduce the operations. Furthermore, Guo \etal~\cite{guo2016dynamic} proposed an on-the-fly connection pruning method named dynamic network surgery, which can avoid incorrect pruning and make it as a continual network maintenance by incorporating connection splicing into the whole process.  The sparse network achieved from the fine-grained pruning methods has much lower computation cost theoretically. Unfortunately, there is no mature framework or hardware for sparse network, and thus only limited speed up can be obtained practically.

	There are also some other methods attempting to do \textit{coarse-grained pruning} by cutting off the entire filters, channels or even layers. In~\cite{he2017channel}, He \etal proposed an iterative two-step algorithm to effectively prune each layer by using a LASSO regression, which based on the channel selection and least square reconstruction. In ~\cite{li2016pruning}, Li \etal applied L1 regularization to prune filters to induce sparsity. More generally, Wen \etal~\cite{wen2016learning} proposed a structured sparsity learning method to reduce redundant filters, channels, and layers in a unified manner. Coarse-grained pruning methods directly remove the filters or channels and thus effectively accelerate the network. 
	
	\textbf{Quantization.}
	Network quantization is to reduce the number of bits used to represent the parameters or features. Han \etal~\cite{chen2015compressing} proposed a deep compression method that firstly pruned the insignificant connections, and secondly quantized the left connections by using weight sharing and Huffman coding. 
	Moreover, INQ~\cite{zhou2017incremental} and ShiftCNN~\cite{gudovskiy2017shiftcnn} quantize a full-precision CNN model to a low-precision model whose parameters (i.e. weights) are either zero or powers of two. With the powers of two representation, the multiplication operations can be replaced by shift operations which is quite efficient.  
	Besides these post-quantization methods, there are also some methods attempting to directly train a binary network, such as BinaryConnect~\cite{courbariaux2015binaryconnect}, BNNs~\cite{courbariaux2016binarized} and XNORNetworks~\cite{rastegari2016xnor}.  As quite fewer bits used, all these methods can obtain faster networks, but correspondingly the accuracy usually is significantly decreased when dealing with large networks.
	
	\textbf{Other Methods.}
	In addition to the methods described above, some other approaches explored how to use low-rank factorization, knowledge distillation 
	for deep network acceleration. In \textit{Low-rank decomposition} methods~\cite{denton2014exploiting, jaderberg2014speeding}, the convolutional filters structured in 4D tensors are decomposed to lower-rank filters, which removes the redundancy in convolution inducing fewer calculations. In \textit{knowledge distillation} methods~\cite{hinton2015distilling, romero2014fitnets},  the knowledge learnt from a deep and wide network is shifted into shallower and narrower ones by making the output distribution of the two networks the same. 
	
	The post-processing methods are simple and intuitive, but obviously have some \textit{limitations.} Most above methods are in two or multiple steps manner,  
	the objective of the network (such as classification or detection) and the objective of acceleration are separately optimized. Therefore, the acceleration does not necessarily ensure excellent classification or detection accuracy.  Besides, most pruning methods determine the importance of a connection or layer by only considering its magnitude and its contribution to several adjacent layers, but not its influence on the whole network. As verified in \cite{yu2017nisp}, pruning without considering the global impact will result in significant error propagation, causing performance degeneration especially in deep networks.
	
	\subsection{Design Efficient Architectures}
	Considering the above mentioned limitations, some researchers go other way to directly design efficient network architectures, such as smaller filters, separable convolution, group convolution, and etc.
	
	\textbf{Separable Convolution.}
	Some early works straightforwardly employ small filters (e.g. 1$\times$1, 3$\times$3) to replace those large ones (e.g. 5$\times$5, 7$\times$7) for acceleration~\cite{simonyan2014very, he2016deep, iandola2016squeezenet, huang2017densely}. However, even if only with 3$\times$3 and 1$\times$1 filters, an ordinary deep network is still time consuming, such as the ResNet50 needs about 4G MAdds\footnote{In this paper, MAdds refers to the number of multiplication-addition operations.} and VGG16 needs 16G MAdds for calculating a $224 \times 224$ image.
	In order to get further acceleration, some works explore separable convolution which uses multiple 2D convolutions to replace the time-intensive 3D convolutions. In aspect of spatial separation, Inception V3~\cite{szegedy2016rethinking} factorizes a $h\times w\times c$ filter into two ones, i.e. one $h\times 1\times c$ filter and another $1\times w\times c$ filter. In aspect of channel separation, Xception~\cite{chollet2017xception} and MobileNets~\cite{howard2017mobilenets} employ depthwise separable convolution. 
	This kind of separable convolution can speed up the computing exponentially, and thus they are widely used in many modern networks.  


	
	\textbf{Group Convolution.}
	Separable convolution achieves the acceleration by factorizing the filters. Differently, the group convolution speed up the network by dividing all filters into several groups, such as ~\cite{ioannou2017deep, sun2018igcv3, xie2018igcv, zhang2017interleaved, xie2017aggregated, zhang1707shufflenet}. The concept of group convolution was first proposed in Alexnet~\cite{krizhevsky2012imagenet}, 
	and then it is further successfully adopted in ResNeXt~\cite{xie2017aggregated}, making it popular in recent network design. However, standard group convolutions do not communicate between different groups, which restricts their representation capability. To solve this problem, in ShuffleNet~\cite{zhang1707shufflenet}, a channel shuffle unit is proposed to randomly permute the output channels of group convolution to make the output better related to the input.  In these methods, the elements (i.e. input channels and filters) in each group are fixed or randomly defined.  Furthermore, in Condensenet~\cite{huang2018condensenet} a learnable group convolution was proposed to automatically select the input channels for each group.
	
	Although the existing group convolution methods have advanced the acceleration very effectively, there are still several limitations to solve: 1) The filters used for group convolution in each group are pre-defined and fixed, and this hard assignment hinders its representation capability even with random permutation after group convolution; 2) In some works the groups are learnable, but usually are designed as a tedious multiple-stage or iterative manner. In each stage, the network from previous stage is firstly pruned and then fine-tuned to recover the accuracy.  

	To deal with all above limitations once for all, in this work we propose a fully learnable group convolution (FLGC) method. In our proposed FLGC, the grouping structure including the input channels and filters in one group is dynamically optimized. What's more, this module can be embedded into any existing deep network and easily optimized in an end-to-end manner.  At test time, the learnt model is calculated similar as the standard group convolution which allows for efficient computation in practice. 
	A brief comparison of different group convolution methods are shown in Figure~\ref{fig:fig1}. Overall, the advantages of our method are as follows: 
	
	(1) The element including input channels and filters in each group are both learnable, allowing for flexible grouping structure and inducing better representation capability;
	
	(2) The group structure in all layers are simultaneously optimized in an end-to-end manner, rather than a multiple-stage or iterative manner (i.e. pruning layer by layer.);
	
	(3) The numbers of input channels and filters in each group are both flexible, while the two numbers must be divisible by the group number in conventional group convolution.

	\section{Fully Learnable Group Convolution(FLGC)}
	
	
	In modern deep networks, the size of filters is mostly $ 3\times3 $ or $ 1\times1 $, and the main computational cost is from the convolution layer.  The $ 3\times3 $ convolutions can be easily accelerated by using the depthwise separable convolution (DSC). And the separation of $ 3\times3 $ convolutions will come along with additional $ 1\times1 $ convolutions. After DSC, the $ 1\times1 $ convolutions contribute the major time-cost, e.g. for a Resnet50 network, after applying DSC to the $ 3\times3 $ convolutions, the computational cost of $ 1\times1 $ convolutions accounts for $83.6\%$ in the whole network. Therefore, how to speed up the $ 1\times1 $ convolution becomes an urgent problem and attracts increasing attentions.
	
	Since the $ 1\times1 $ filters are non-separable, group convolution becomes a hopeful and feasible solution. 
	However, simply applying group convolution will result in drastic precision degradation. As analyzed in \cite{huang2018condensenet}, this is caused by the fact that the input channels to the $ 1\times1 $ convolutional layer have an intrinsic order or they are far more diverse. This implies that the hard assignment grouping mechanism in standard group convolution is incompetent. For a better solution, our proposed method dynamically determines the input channels and filters for each group, forming a flexible and efficient grouping mechanism.
	
	Briefly, in our FLGC the input channels and filters in one group (i.e. the group structure) are both dynamically determined and updated according to the gradient of the overall loss of the network through back propagation. And thus it can be optimized in an end-to-end manner.

	\subsection{Method}
	In a deep network, the convolution layer is computed as convolving the input feature maps with filters.  Taking the $k^{th}$ layer for an example, the input of the $k^{th}$ layer can be denoted as $X^k = \{x_1^k, x_2^k,\cdots, x_C^k\}$, where $C$ is the number of channels and $x_i^k$ is the $i^{th}$ feature map. The filters of the $k^{th}$ layer are denoted as $W^k = \{w_1^k, w_2^k, \cdots, w_N^k\}$,  where $N$ denotes the number of filters, i.e. number of output channels, and $w_i^k$ is the $i^{th}$ 3D convolutional filter. The output\footnote{We omit the bias $b$ for simplicity.} of this convolution layer is calculated as follows:
	\begin{eqnarray}
	X^{k+1} &=& W^k\otimes X^k\notag \\ 
	&=& \{w_1^k*X^k, w_2^k*X^k, \cdots, w_N^k*X^k\},
	\label{Equation:Eq1}
	\end{eqnarray}
	where $ \otimes $ in this work denotes the convolution between two sets, $*$ denotes the convolution operation between a filter and the input feature maps.
	
	In group convolution, the input channels and filters are divided into $G$ groups respectively, denoted as $X^k = \{X_1^k, X_2^k,\cdots, X_G^k\}$\footnote{$X_1^k\cup X_2^k\cup \cdots\cup X_G^k = X^k$} and  $W^k = \{W_1^k, W_2^k,\cdots, W_G^k\}$\footnote{$W_1^k\cup W_2^k\cup \cdots\cup W_G^k = W^k$}. Now, $X^{k+1}$ is reformulated as below:
	\begin{eqnarray}
	X^{k+1}= \{W_1^k\otimes X_1^k, W_2^k\otimes X_2^k, \cdots, W_G^k\otimes X_G^k\}.
	\label{Equation:Eq2}
	\end{eqnarray}
	
	Typically, in standard group convolution the input channels and filters are evenly divided into $G$ groups in a hard assignment manner, i.e. $\frac{C}{G}$ input channels and $\frac{N}{G}$ filters in each group. Therefore, the number of channels used in each filter is reduced to $\frac{1}{G}$ of original ones, resulting in a acceleration rate as below:
	\begin{eqnarray}
	\frac{\textit{MAdds}(W^k\otimes X^k)}{\textit{MAdds}(\sum_{i=1}^{G}W_i^k\otimes X_i^k)} = G.
	\label{Equation:Eq3}
	\end{eqnarray} 
	
	As can be seen, this group convolution from hard assignment can easily bring about considerable acceleration of $G\times$. However, it is not necessarily a promising approach for accuracy. Therefore, \textit{the goal of our method is to design a fully learnable grouping mechanism, where the grouping structure is dynamically optimized for favorable acceleration as well as accuracy.}
	
	Firstly, we formulate the grouping structure in the $k^{th}$ layer as two binary selection matrices for input channels and filters respectively, denoting as  $S^k$  and $T^k$.

	
	The $S^k$  is a matrix for channel selection in shape of $C\times G$,  with each element defined as:
	\begin{eqnarray}
	S^k(i,j) = 
	\left\{
	\begin{array}{lr}
	1, \text{\,\,if\,\,}  x_i^k \in X_j^k, &  \\
	&  i=[1,C]; j\in[1,G].\\
	0, \text{\,\,if\,\,}  x_i^k \notin X_j^k, &  
	\end{array}
	\right.
	\label{Equation:Eq4}
	\end{eqnarray} 
	in which $S^k(i,j) = 1$ means the  $i^{th}$ input channel is selected into the $j^{th}$ group. As can be seen, the $j^{th}$ column of $S^k$ indicates which input channels belong to the $j^{th}$ group. Then, $X_j^k$ can be simply represented as follows:
	\begin{eqnarray}
	X_j^k = X^k\odot S^k(:,j)^\mathcal{T}, j\in [1,G],
	\label{Equation:Eq5}
	\end{eqnarray} 
	where $\odot$ denotes the element-wise selection operator and the element here means $ \forall x_i^k \in X^k$, and $\mathcal{T}$ denotes the transpose of a vector.
	
	Similarly, for filter selection we define a  matrix $T^k$ in shape of $N\times G$ , with each element defined as:
	\begin{eqnarray}
	T^k(i,j) = 
	\left\{
	\begin{array}{lr}
	1, \text{\,\,if\,\,}  w_i^k \in W_j^k, &  \\
	&  i=[1,N]; j\in[1,G].\\
	0, \text{\,\,if\,\,}  w_i^k \notin W_j^k, &  
	\end{array}
	\right.
	\label{Equation:Eq6}
	\end{eqnarray} 
	in which $T^k(i,j) = 1$ means the $i^{th}$ filter is selected into the $j^{th}$ group. The $j^{th}$ column of $T^k$ indicates which filters belong to the $j^{th}$ group. Then the $j^{th}$ group of filters, i.e. $W_j^k$,  can be represented as:
	\begin{eqnarray}
	W_j^k = W^k\odot T^k(:,j)^\mathcal{T}, j\in [1,G].
	\label{Equation:Eq7}
	\end{eqnarray}  
	
	%
	As a result, the overall group convolution in Eq.(\ref{Equation:Eq2}) can be re-formulated as follows:
	\begin{align}
	X^{k+1}=&W^k\otimes X^k \notag\\
	=& \{W_1^k\otimes X_1^k,W_2^k\otimes X_2^k, \cdots, W_G^k\otimes X_G^k\} \notag\\
	=& \{W^k\odot T^k(:,1)^\mathcal{T} \otimes X^k\odot S^k(:,1)^\mathcal{T} , \cdots, \notag\\
	& W^k\odot T^k(:,G)^\mathcal{T} \otimes X^k\odot S^k(:,G)^\mathcal{T}\}.
	\label{Equation:Eq8}
	\end{align} 
	
	
	With Eq.(\ref{Equation:Eq8}), the structure of group convolution is parameterized by two binary selection matrix $S^k$ and $T^k$. Therefore, this parameterized group convolution can be embedded in any existing deep networks with the objective as:
	\begin{eqnarray}
	\min\limits_{W^k,S^k,T^k|_{k=1}^K}\frac{1}{n}\sum_{i=1}^{n}L(Y_i;\hat{Y}|X_i,W^k,S^k,T^k),
	\label{Equation:Eq9}
	\end{eqnarray}
	in which $X_i$ denotes the $i^{th}$ input sample, $n$ indicates the number of training data, $Y_i$ indicates the $i^{th}$ sample's true category label, $K$ is the number of layers, and $\hat{Y}$ is the label predicted from a network with our group convolution parameterized by $W^k,S^k,T^k$.  $L(,)$ denotes the loss function (e.g. cross entropy loss) for classification or detection etc.
	
	In the above objective, the filters $W^k$, the group structure including $S^k$ and $T^k$ can be all automatically optimized according to the overall objective function. However, binary variables are notorious for its non-differential feature. So, we further design an ingenious approximation to make it differentiable for better optimization in section \ref{sec: sec2.2}.  
	
	As can be seen from Eq.(\ref{Equation:Eq9}), the group structure in our method is automatically optimized rather than manually defined. 
	Furthermore, different from those methods only considering the magnitude and impact of the connection in one or two layers, the group structure in our method  is determined  according to the objective loss of the whole network. Therefore, the group structures of all layers in our method are jointly optimized implying a superior solution.

	\subsection{Optimization}
	\label{sec: sec2.2}
	In Eq.(\ref{Equation:Eq9}), the filters $W^k$ can be easily optimized as most deep networks by using the stochastic gradient descent, while the binary parameters are hard to optimize as they are non-differentiable. To solve this problem, we approximate the $ {S}^k$ and $ {T}^k$ by applying a softmax function to a meta selection matrix to make it differentiable.
	
	Specifically, we introduce a meta selection matrix $\bar{S}^k$ for channel selection with the same shape as $S^k$. And then the softmax function is applied to each row of $\bar{S}^k$, which can map it to (0,1) as below:
	\begin{eqnarray}
	\hat{S}^k(i,:)&=&softmax(\bar{S}^k(i,:)), \,i\in [1,C].
	\label{Equation:Eq10}
	\end{eqnarray}
	Here, the meta selection matrix $\bar{S}^k$ can be initialized as Gaussian distribution or results from other methods. After softmax, the $i^{th}$ row of $\hat{S}^k$ indicates the probability that the $i^{th}$ input channel belongs to each group. So, the $i^{th}$ input channel can be simply selected into the group with highest probability. That is to say, the binary selection matrix $S^k$ can be approximated as:
	\begin{eqnarray}
	S^k(i,j) = 
	\left\{
	\begin{array}{lr}
	1, \text{\,\,if\,\,}  \hat{S}^k(i,j) = max(\hat{S}^k(i,:)), &  \\
	& \\
	0, \text{\,\,otherwise}  . &  
	\end{array}
	\right.
	\label{Equation:Eq11}
	\end{eqnarray} 
	
	The reason of using softmax function is that with softmax operation the meta selection matrix $\bar{S}^k$ can be updated  to make the output $\hat{S}^k$ approximating 0 or 1 as close as possible. As a result, the quantization error between $\bar{S}^k$ and $S^k$ is largely narrowed.  

	Similarly, the binary selection matrix $T^k$ is approximated by applying softmax function on a meta selection matrix $\bar{T}^k$ for filter selection as follows:
	\begin{eqnarray}
	T^k(i,j) = 
	\left\{
	\begin{array}{lr}
	1, \text{\,\,if\,\,}  \hat{T}^k(i,j) = max(\hat{T}^k(i,:)), &  \\
	& \\
	0, \text{\,\,otherwise}  , &  
	\end{array}
	\right.
	\label{Equation:Eq12}
	\end{eqnarray} 
	with 
	\begin{eqnarray}
	\hat{T}^k(i,:)&=&softmax(\bar{T}^k(i,:)), \,i\in [1,N].
	\label{Equation:Eq13}
	\end{eqnarray}
	Here, the $i^{th}$ row of $\hat{T}^k$ indicates the probability that the $i^{th}$ filter belongs to each group.

	In summary,  with the above Eq.(\ref{Equation:Eq10}), Eq.(\ref{Equation:Eq11}), Eq.(\ref{Equation:Eq12})and Eq.(\ref{Equation:Eq13}), the differentiation of the binary $S^k$ and $T^k$ are shifted to the differentiation of the meta selection variable $\bar{S}^k$ and $\bar{T}^k$ which are non-binary, yet with small quantization error.
	
	Furthermore, for easy implementation, Eq.(\ref{Equation:Eq8}) is equivalently transformed to the following formulation:
	\begin{align}
	X^{k+1}=& \{W^k\odot T^k(:,1)^\mathcal{T} \otimes X^k\odot S^k(:,1)^\mathcal{T} , \cdots, \notag\\
	& W^k\odot T^k(:,G)^\mathcal{T} \otimes X^k\odot S^k(:,G)^\mathcal{T}\} \notag\\
	=& (W^k\odot M^k) \otimes X^k,
	\label{Equation:Eq14}
	\end{align} 
	with $M^k = T^k(S^k)^\mathcal{T}$ , and the shape of $M^k$ is $N\times C$ that is the same as $W^k$.

	Finally, the objective function is re-formulated as below:
	\begin{eqnarray}
	\min\limits_{W,\bar{S},\bar{T}}\frac{1}{n}\sum_{i=1}^{n}L(Y_i, X_i(W \odot M)+b),
	\label{Equation:Eq15}
	\end{eqnarray}
	where $W=\{W^k|_{k=1}^K\}$, $\bar{S}=\{\bar{S}^k|_{k=1}^K\}$, $\bar{T}=\{\bar{T}^k|_{k=1}^K\}$.
	
	The objective in Eq.(\ref{Equation:Eq15}) can be easily optimized as most deep network by using the stochastic gradient descent method, with the parameters of each layer are updated as follows:
	%
	%
	%
	%
	%
	%
	%
	%
	
	\begin{eqnarray}
	W^k_{(i,j)} \gets W^k_{(i,j)} - \eta \frac{\partial L}{\partial\left(W^k_{(i,j)}\odot M^k_{(i,j)}\right)} , \forall i,j \in I,
	\label{Equation:Eq16}
	\end{eqnarray}
	
	\begin{align}
	\bar{S}^k_{(i,j)} \gets &\bar{S}^k_{(i,j)} - \eta \frac{\partial L}{\partial\left(W^k_{(i,j)}\odot M^k_{(i,j)}\right)} \notag \\
	&\frac{\partial\left(W^k_{(i,j)}\odot M^k_{(i,j)}\right)}{\partial M^k_{(i,j)}}  \frac{\partial M^k_{(i,j)}}{\partial \hat{S}^k_{(i,j)}} \frac{\partial \hat{S}^k_{(i,j)}}{\partial \bar{S}^k_{(i,j)}},
	\label{Equation:Eq17}
	\end{align}
	
	\begin{align}
	\bar{T}^k_{(i,j)} \gets &\bar{T}^k_{(i,j)} - \eta \frac{\partial L}{\partial\left(W^k_{(i,j)}\odot M^k_{(i,j)}\right)} \notag \\
	&\frac{\partial\left(W^k_{(i,j)}\odot M^k_{(i,j)}\right)}{\partial M^k_{(i,j)}} \frac{\partial M^k_{(i,j)}}{\partial \hat{T}^k_{(i,j)}} \frac{\partial \hat{T}^k_{(i,j)}}{\partial \bar{T}^k_{(i,j)}},
	\label{Equation:Eq18}
	\end{align}
	in which $\eta$ indicates the learning rate. The overall procedure is summarized in Algorithm~\ref{algorithm:alg1}.

	\begin{algorithm}[h] 
		\caption{FLGC: solving the optimization problem in Eq.(\ref{Equation:Eq15}) via SGD}  
		\label{algorithm:alg1}
		\begin{algorithmic}[1]
			\Require X: training data, Y: lable
			\Ensure $\{W^k, S^k, T^k: k\in[1,K]\} $ 
			\State Initialize $W^k \gets msra$; $\bar{S}^k, \bar{T}^k \gets Gaussian$; $S^k, T^k \gets 0$
			\For {each batch $X_i$}
			\State //Forward propagation:
				\For{$i = 1 \to C$}
				\State$\hat{S}^k(i,:) \gets softmax(\bar{S}^k(i,:))$
				\State $S^k(i,j) \gets 1$ ,if $\hat{S}^k(i,j) = max(\hat{S}^k(i,:))$
				\EndFor
				\For{$i = 1 \to N$}
				\State$\hat{T}^k(i,:)\gets softmax(\bar{T}^k(i,:))$
				\State $T^k(i,j) \gets 1$ ,if $\hat{T}^k(i,j) = max(\hat{T}^k(i,:))$
				\EndFor
				\State $M^k\gets T^k(S^k)^T$
				\State Get loss: $L = L(Y_i, X_i(W^k \odot M^k)+b)$
			
			\State //Backward propagation:
				\State $W^k \gets W^k - \eta \frac{\partial L}{\partial(W^k\odot M^k)}$
				\State $\bar{S}^k \gets \bar{S}^k - \eta \frac{\partial L}{\partial(W^k\odot M^k)}\frac{\partial(W^k\odot M^k)}{\partial M^k}  \frac{\partial M^k}{\partial \hat{S}^k} \frac{\partial \hat{S}^k}{\partial \bar{S}^k}$
				\State $\bar{T}^k \gets \bar{T}^k - \eta \frac{\partial L}{\partial(W^k\odot M^k)} \frac{\partial(W^k\odot M^k)}{\partial M^k}  \frac{\partial M^k}{\partial \hat{T}^k} \frac{\partial \hat{T}^k}{\partial \bar{T}^k}$
			\EndFor
		\end{algorithmic}  
	\end{algorithm}

	\subsection{Inference with Index-Reordering}
	\begin{figure}[hb]
		\begin{center}
			\includegraphics[width=1\linewidth]{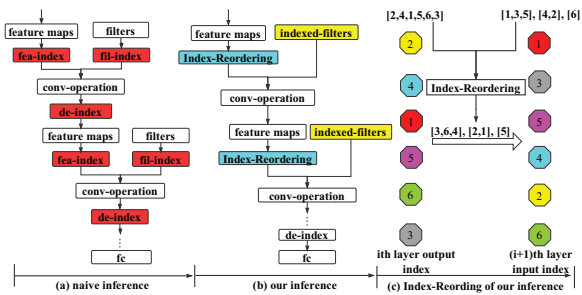}
		\end{center}
		\caption{Illustration of index re-ordering for efficient inference. (a) is a naive inference method, (b) is our efficient inference method, (c) is Index-Reordering unit.}
		\label{fig:fig2}
		\vspace{-10pt}
	\end{figure}
	After the group structure is learnt, the input channels and filters usually need to be re-organized for fast inference. A naive method is to add an index layer to re-order the input channels according to the group information, and another index layer to re-order the filters. Then, the output channels are re-ordered back to the original order, as shown in Figure~\ref{fig:fig2}(a). Unfortunately, such frequent re-order operations on memory will significantly increase the inference time. 
	
	Therefore, we propose an efficient strategy for index re-ordering as shown in Figure~\ref{fig:fig2}(b). Firstly, the filters are re-ordered to make those filters in one group arranged together. Secondly, considering that the input channels are also the output channels of previous layer, we merge the index of the output from previous layer and index of input channels in this layer as single index to obtain correct order of input channels. The detail is shown in Figure~\ref{fig:fig2}(c). 
	As designed like above, the operations on memory are reduced a lot and all these re-ordering index can be obtained offline, so it is quite efficient at the inference stage. 
	
	As a result, at the inference time our FLGC can be as efficient as the standard group convolution. 
	
	\section{Experiments}
	\begin{figure*}[h]
		\begin{center}
			\includegraphics[width=1\linewidth]{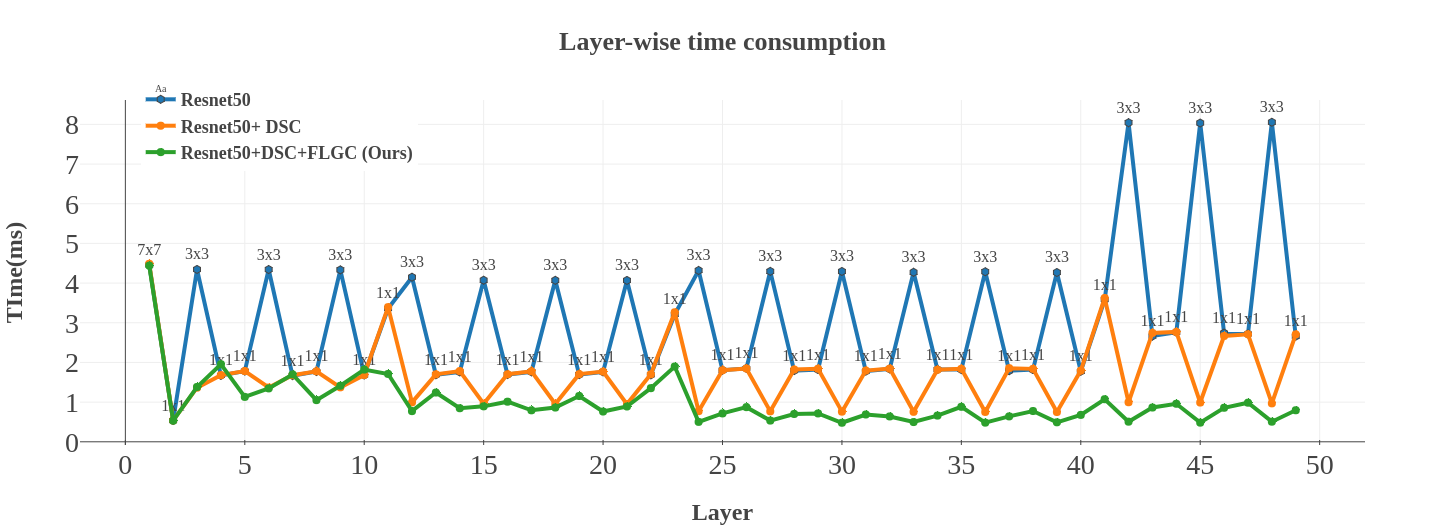}
		\end{center}
		\caption{The time cost of each convolutional layer in Resnet50 with different convolution mechanisms on a single CPU. The blue line is the Standard Resnet50. The orange line is the Resnet50 with $ 3\times3 $ convolutions replaced by DSC. The green line is the Resnet50 with $ 1\times1 $ convolutions further replaced by FLGC.}
		\label{fig:fig4}
		\vspace{-10pt}
	\end{figure*}
	In this section, we investigate the effectiveness of our proposed FLGC by embedding it into the existing popular CNNs networks including Resnet50~\cite{he2016identity}, MobileNetV2~\cite{sandler2018inverted} and Condensenet~\cite{huang2018condensenet}. Firstly, we conduct ablation study of FLGC on  CASIA-WebFace~\cite{yi2014learning}, and then compare it with existing competitive methods on CASIA-WebFace, CIFAR-10 and ImageNet (ILSVRC 2012)~\cite{deng2009imagenet} in terms of face verification and image classification. 
	
	\subsection{Embedding into the state-of-the-art CNNs}
	\label{sec: sec3.1}
	We select three state-of-the-art architectures including Resnet50, MobileNetV2 and CondenseNet to embed the proposed fully learnable group convolution(FLGC) for evaluation. 
	
	\textbf{Resnet50 with FLGC.} 
	The Resnet50 is a powerful network which achieves prominent accuracy on many tasks. Nevertheless, it is quite time-consuming. As shown in Figure~\ref{fig:fig4}(blue line),  the major computation cost falls on the $ 3\times3 $ convolutions, and thus we firstly use the DSC to separate the $ 3\times3 $ convolutions following MobileNet~\cite{howard2017mobilenets}. After DSC, there are a large number of $ 1\times1 $ convolutions, which computational cost accounts for $83.6\%$ of the whole network. Therefore, we further replace all the $ 1\times1 $ layers in the network with our FLGC layers. 
	Besides, we simply double the stride of the first layer and add a fc layer. 
	
	\textbf{MobileNetV2 with FLGC.} 
	The MobileNetV2 is a state-of-the-art efficient architecture with elaborative design. This architecture achieves satisfactory accuracy on many tasks with favorable computational cost, e.g. classification, detection and segmentation. But still, the intensive $ 1\times1 $ convolutions takes great majority of computational cost, leaving much room for further acceleration. Therefore, we  replace those $ 1\times1 $ convolution layers, of which the filters number is larger than 96, with our FLGC layer. 
	
	\textbf{CondenseNet with FLGC.} 
	CondenseNet proposed a learnable group convolution which can automatically select the input channels for each group. However, the filters in each group are fixed, and the process are designed as a tedious multiple-stage or iterative manner. Besides, the importance of each input channel is determined according to the magnitude of the connections between the input and the filters, but without considering its impact on the overall network, leading to a sub-optimal solution. We substitute all the FLGC for the LGC in CondenseNet.

	\subsection{Ablation Study}
	The ablation experiment is conducted on CASIA-WebFace with Resnet50 in terms of face verification. The experimental setting on this dataset is the same as that in section \ref{sec: sec3.1}. 
	
	Firstly, we analyze the speedup with DSC and our FGLC by comparing with the standard convolution. The time cost of each layer in all methods are shown in Figure~\ref{fig:fig4}. As can be seen, in the standard Resnet50 denoted in the blue line, $ 3\times3 $ convolution layer is the major time-consuming part. After applying DSC, $ 3\times3 $ convolution time cost is significantly reduced as shown in the orange line, and the orange line also highlights that $ 1\times1 $ convolution layer is the major time cost part now. By further applying FLGC, the time cost of $ 1\times1 $ convolution layer is successfully reduced as shown in the green line, resulting in a quite efficient architecture with comparable accuracy as the baseline(standard Resnet50). For overall procedure, our method achieves a significant improvement of time cost.
	\begin{figure}[h]
		\begin{center}
			\includegraphics[width=1\linewidth]{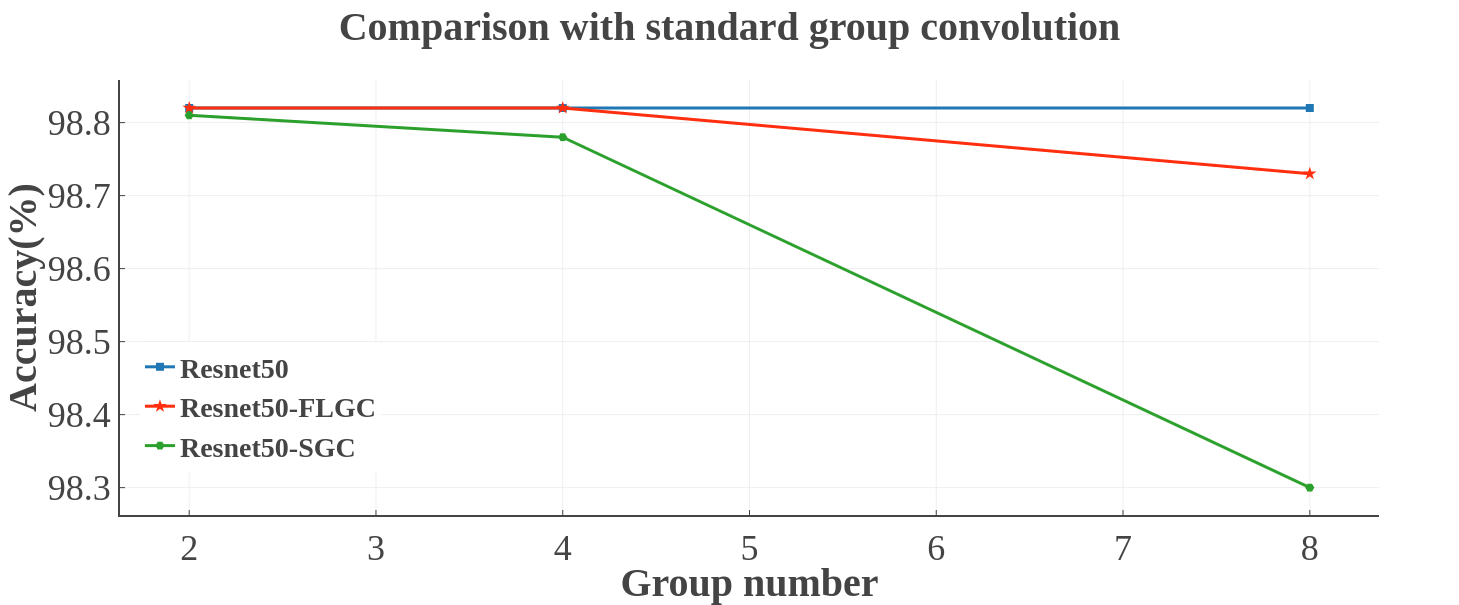}
		\end{center}
		\caption{Compare our FLGC with standard group convolution(SGC) in terms of face verification accuracy of Resnet50 on CASIA-WebFace w.r.t. different group numbers.}
		\label{fig:fig6}
		\vspace{-10pt}
	\end{figure}
	
	Besides the efficiency, we further explore the accuracy of standard group convolution and our FLGC w.r.t. different number of groups, and the result are shown in Figure~\ref{fig:fig6}, Table~\ref{table:table1} and Table~\ref{table:table2}. As can be seen, the accuracy drops dramatically when the standard group convolution is applied to the $ 1\times1 $ convolution, mainly due to the loss of representation capability from hard assignment. Differently, our FLGC successfully maintains the accuracy even with large number of groups, benefitted from the fully learnable mechanism for grouping structure. 
	
	
	\subsection{Comparison with competitive approaches}
    \textbf{Results on CASIA-WebFace.}
	The CASIA-WebFace is a commonly used wild dataset for face verification, consisting of 494,414 face images from 10,575 subjects. All faces are detected and aligned by using \cite{wu2017funnel} and \cite{zhang2016occlusion}, and then the detected faces are cropped out in resolution of 256$\times$256. This dataset is used for training. Following the mainstream works, the well-known LFW~\cite{huang2008labeled} dataset is used for face verification evaluation. LFW includes 13,233 face images from 5749 different identities, and the standard protocol defines 3,000 positive pairs and 3,000 negative pairs for verification testing.

	On this dataset, we embed the proposed FLGC into the Resnet50 as described in section 3.1. For optimization of our method, we initialize the meta selection matrix $\bar{S}^k$ and $\bar{T}^k$ with Gaussian distribution, and simply set the hyperparameters of momentum as 0.9, weight decay as $ 5\times10^{-4} $, batch size as 80, and iterations as 120,000. Two versions of our FLGC with group number as 4 and 8 are evaluated respectively.  
	
	
	Our accelerated network is compared with several state-of-the-art methods on this dataset including \cite{yi2014learning, liu2017sphereface, ding2015robust, liu2016large}. All methods for comparison including ours employ softmax loss for optimization.  The experimental results are shown in Table~\ref{table:table1}. As can be seen, the standard Resnet50 achieves better verification rate with giant architecture than \cite{yi2014learning, liu2017sphereface, ding2015robust, liu2016large},  inevitably leading to high computational cost. Expectedly, our modified Resnet50 achieves about 18x speed up over standard Resnet50 without accuracy drop, which is also much faster than \cite{yi2014learning, liu2017sphereface, ding2015robust, liu2016large}. In practical evaluation on single CPU(Intel(R) Xeon(R) CPU E5-2620 v3 @2.40GHz), our modified Resnet50 runs 5x faster than standard Resnet50, demonstrating the effectiveness of our method.
	
	\begin{table}
		\caption{Face verification accuracy (\%) and time complexity on LFW, all the medels are trained on CASIA-WebFace. The archtecture of ResNet50-FLGC and ResNet50-SGC are introduced in Section~\ref{sec: sec3.1}. (SGC: standard group convolution) }
		\begin{tabularx}{8cm}{lXlXl}
			\toprule[1.5pt]
			\textbf{Model} & \textbf{MAdds}  & \textbf{Params} & \textbf{Acc} \\
			\midrule[1.5pt]
			Yi \etal~\cite{yi2014learning}  & 770M & 1.75M & 97.73 \\
			64layer+Softmax~\cite{liu2017sphereface} & 28460M & 37.16M & 97.88 \\
			Ding \etal~\cite{ding2015robust}  & 2874M & 3.76M & 98.43 \\
			Liu \etal~\cite{liu2016large}  & 10194M & 6.78M & 98.71 \\
			\hline
			ResNet50(stardand)  & 3727M & 20.69M & 98.82 \\
			ResNet50-SGC(G=2)  & 363M & 5.35M & 98.81 \\
			\textbf{ResNet50-FLGC(G=2)}  & \textbf{363M} & \textbf{5.35M} & \textbf{98.82} \\
			\hline
			ResNet50-SGC(G=4)  & 203M & 2.70M & 98.78 \\
			\textbf{ResNet50-FLGC(G=4)}  & \textbf{203M} & \textbf{2.70M} & \textbf{98.82} \\
			\hline
			ResNet50-SGC(G=8)  & 124M & 1.37M & 98.30 \\
			\textbf{ResNet50-FLGC(G=8)}  & \textbf{124M} & \textbf{1.37M} & \textbf{98.73} \\
			\bottomrule[1.5pt]
		\end{tabularx}
		\label{table:table1}
		\vspace{-10pt}
	\end{table}

	%
	

	\textbf{Results on CIFAR-10.}
	We further compare our FLGC with other acceleration approaches on CIFAR-10 dataset. CIFAR-10 consists of 10 classes and 60,000 images in resolution of 32$\times$32 pixels. Among them, 50,000 images are used for training and 10,000 for testing. 
	
	Since the image resolution on this dataset is small, the modified Resnet50 in Section~\ref{sec: sec3.1} used for 224$\times$224 image is too large and redundant. So, we replace the 7$\times$7 convolution layer with 3$\times$3 convolution layer to suit the smaller input images. Based on this baseline architecture, we replace the 1x1 convolution layers with FLGC layers, and the number of group is set as 4. For clear comparison, two versions of FLGC with different MAdds by changing number of filters is proposed, referred to as ResNet50-FLGC1 and ResNet50-FLGC2.  Besides Resnet50, we also embed our FLGC in the state-of-the-art acceleration architecture MobileNetV2, referred to as MobileNetV2-FLGC. For optimization of our method, all hyperparameters is the same as that used on CASIA-WebFace.
	
	On this dataset, we compare the FLGC with state-of-the-art filter level pruning methods and the state-of-the-art architecture MobileNetV2. The comparison results are shown in Table~\ref{table:table2}. Comparing with the pruning methods~\cite{he2017channel, li2016pruning} which also employ the Resnet architecture, we can get lower classification error with 3$\times$ fewer MAdds. Besides, our FLGC can be flexibly embedded into any efficient architectures such as MobileNetV2, leading to further speedup. As can be seen in Table~\ref{table:table2}, MobileNetV2 with FLGC achieves better accuracy w.r.t different group number, further demonstrating the superiority of our proposed FLGC.

	\begin{table}[h]
		\caption{Image classification error(\%) and time complexity of different methods on CIFAR-10.(G:group number)}
		\begin{tabularx}{8.2cm}{p{4cm}p{1cm}p{1cm}p{0.7cm}}
			\toprule[1.5pt]
			\textbf{Model} & \textbf{MAdds}  & \textbf{Params} & \textbf{Err} \\
			\midrule[1.5pt]
			ResNet56-pruned~\cite{he2017channel}  & 62M & --- & 8.2 \\
			\textbf{ResNet50-FLGC1(ours)}  & \textbf{23M} & \textbf{0.22M} & \textbf{7.95} \\
			\hline
			ResNet56-pruned~\cite{li2016pruning}  & 90M & 0.73M & 6.94 \\
			\textbf{ResNet50-FLGC2(ours)}  & \textbf{44M} & \textbf{0.68M} & \textbf{6.77} \\
			\hline
			MobileNetV2-SGC(G=2)  & 158M & 1.18M & 6.04 \\
			\textbf{MobileNetV2-FLGC(G=2)}  & 158M & 1.18M & \textbf{5.89} \\
			\textbf{MobileNetV2-FLGC(G=3)}  & 122M & 0.85M & \textbf{5.80} \\
			MobileNetV2-SGC(G=4)  & 103M & 0.68M & 6.64 \\
			\textbf{MobileNetV2-FLGC(G=4)}  & 103M & 0.68M & \textbf{5.84} \\
			\textbf{MobileNetV2-FLGC(G=5)}  & 92M & 0.58M & \textbf{6.12} \\
			\textbf{MobileNetV2-FLGC(G=6)}  & 85M & 0.51M & \textbf{6.33} \\
			\textbf{MobileNetV2-FLGC(G=7)}  & 80M & 0.46M & \textbf{6.34} \\
			MobileNetV2-SGC(G=8)  & 76M & 0.43M & 7.51 \\
			\textbf{MobileNetV2-FLGC(G=8)}  & 76M & 0.43M & \textbf{6.91} \\
			\bottomrule[1.5pt]
		\end{tabularx}
		\label{table:table2}
		\vspace{-10pt}
	\end{table}

	\textbf{Results on ImageNet.}
	To further validate the effectiveness of our proposed FLGC, we compare our FLGC with state-of-the-art learnable group convolution which proposed in CondenseNet~\cite{huang2018condensenet} on ImageNet.  
	
	For a fair comparison, we used the same network structure as CondenseNet. Based on this baseline architecture, we replace the LGC layers in CondenseNet with our FLGC layers and standard group convolution (SGC) layers respectively, and the number of group is set as 4. What's more, we keep the hyperparameters the same as that used in CondenseNet. All models are trained for 120 epochs, with a cosine shape learning rate which starts from 0.2 and gradually reduces to 0. As can be seen in Table~\ref{table:table3}, our FLGC achieves better accuracy than CondenseNet's LGC and SGC. Moreover, Our FLGC even achieves a favorable performance compared with competitive MobileNet, ShuffleNet and NASNet-A.
	
	\begin{table}
		\caption{Comparison of Top-1 and Top-5 classification error rate
			(\%) with other state-of-the-art compact models on ImageNet. }
		\begin{tabularx}{8.3cm}{p{3.5cm}p{0.8cm}p{0.8cm}p{0.5cm}p{0.5cm}}
			\toprule[1.5pt]
			\textbf{Model} & \textbf{MAdds}  & \textbf{Params} & \textbf{Top1} & \textbf{Top5}  \\
			\midrule[1.5pt]
			Inception V1\cite{szegedy2015going}  & 1448M & 6.6M & 30.2 &10.1 \\
			1.0 MobileNet-224\cite{howard2017mobilenets} & 569M & 4.2M & 29.4 & 10.5 \\
			ShuffleNet 2x\cite{zhang1707shufflenet}  & 524M & 5.3M & 26.3 & --- \\
			NASNet-A (N=4)\cite{zoph2018learning}  & 564M & 5.3M & 26.0 & 8.4 \\
			NASNet-B (N=4)\cite{zoph2018learning}  & 488M & 5.3M & 27.2 & 8.7 \\
			NASNet-C (N=4)\cite{zoph2018learning}  & 558M & 4.9M & 27.5 & 9.0 \\
			CondenseNet (G=4)\cite{huang2018condensenet}  & 529M & 4.8M & 26.2 & 8.3 \\
			CondenseNet-SGC  & 529M & 4.8M & 29.0 & 9.9 \\
			\midrule
			\textbf{CondenseNet-FLGC}  & \textbf{529M} & \textbf{4.8M} & \textbf{25.3} & \textbf{7.9} \\
			\bottomrule[1.5pt]
		\end{tabularx}
		\label{table:table3}
		\vspace{-10pt}
	\end{table} 

	\section{Conclusion}
	In this work, we propose a fully learnable group convolution module which is quite efficient and can be embedded into any layer of any deep neural networks for acceleration. Instead of the existing pre-defined, two-steps, or iterative acceleration strategies, FLGC can automatically learn the group structure at the training stage according to the overall loss of the network in a fully end-to-end manner, and run as efficient as standard group convolution at the inference stage. The number of input channels and filters in each group are flexible, which ensures better representation capability and well solves the problem of uneven information distribution encountered in standard group convolution. 
	Furthermore, compared with LGC of CondenseNet and standard group convolution, our FLGC can better maintain the accuracy while achieve significant acceleration even with large number of groups. 
	
	\section*{Acknowledgements}
	This work is partially supported by the National Key R\&D Program of China (No. 2017YFA0700800), Natural Science Foundation of China (Nos. 61650202, 61772496 and 61532018).
	
    \newpage
	
	{\small
		\bibliographystyle{ieee}
		\bibliography{egbib}
	}
	
\end{document}